\documentclass{article}

\usepackage[nonatbib,preprint]{neurips_wrl2022}

\usepackage[utf8]{inputenc} 
\usepackage[T1]{fontenc}    
\usepackage{hyperref}       
\usepackage{url}            
\usepackage{booktabs}       
\usepackage{amsfonts}       
\usepackage{nicefrac}       
\usepackage{microtype}      
\usepackage{graphicx}
\usepackage{cite}
\title{Knolling bot 2.0: Enhancing Object Organization with Self-supervised Graspability Estimation}

\author{%
    Yuhang Hu \\
    Department of Mechanical Engineering\\
    Columbia University\\
    New York, NY 10027 \\
    \texttt{yuhang.hu@columbia.edu} \\
    \And
    Zhizhuo Zhang \\
    Department of Mechanical Engineering\\
    Columbia University\\
    New York, NY 10027 \\
    \texttt{zz3012@columbia.edu} \\
    \And
    Hod Lipson \\
    Data Science Institute, Department of Mechanical Engineering\\
    Columbia University\\
    New York, NY 10027 \\
    \texttt{hod.lipson@columbia.edu} 
}

\begin{document}

\maketitle

\begin{abstract}
Building on recent advancements in transformer-based approaches for domestic robots performing 'knolling'—the art of organizing scattered items into neat arrangements—this paper introduces Knolling bot 2.0. Recognizing the challenges posed by piles of objects or items situated closely together, this upgraded system incorporates a self-supervised graspability estimation model. If objects are deemed ungraspable, an additional behavior will be executed to separate the objects before knolling the table. By integrating this grasp prediction mechanism with existing visual perception and transformer-based knolling models, an advanced system capable of decluttering and organizing even more complex and densely populated table settings is demonstrated. Experimental evaluations demonstrate the effectiveness of this module, yielding a graspability prediction accuracy of 95.7\%.
\end{abstract}

\begin{figure}
  \centering
  \includegraphics[width=0.8\textwidth]{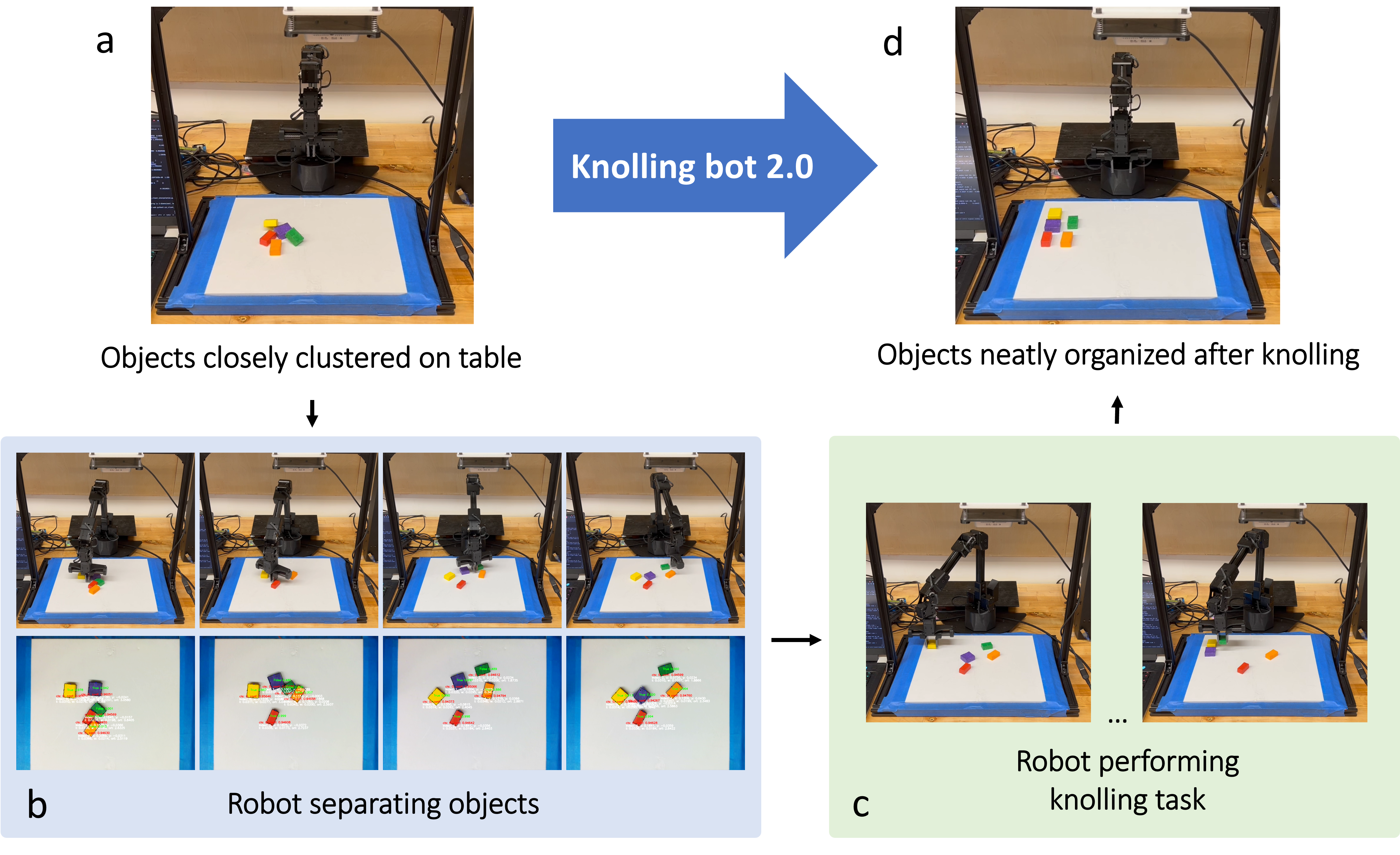}
  \caption{\textbf{Knolling bot 2.0 Pipeline.}(a) Five colored objects closely clustered on the table. (b) Top row: Four images depicting the robot arm in the process of separating objects. Bottom row: Corresponding top-down camera views for each image above, annotated with outputs from the visual perception and graspability estimation models. (c) Two images show the robot performing the knolling task. (d) Final arrangement post-knolling: The robot has returned to its origin, and the objects are neatly organized on the front left of the table. For a comprehensive view of this demonstration, refer to the supplementary movie1.}
\end{figure}

\section{Introduction}
Domestic robots are increasingly being recognized as valuable aids for routine human tasks. Among these, a critical manipulation task is organizing and tidying tables.\cite{kapelyukh2023dall,liu2022structformer,xu2023tidy,zeng2021transporter,manuelli2019kpam,goodwin2022semantically,danielczuk2021object}. Unlike structured industrial settings, household environments are dynamic, with a diverse array of items and ever-changing configurations\cite{lee2021service, kim2019control,abdo2015robot}. Furthermore, human preferences play a pivotal role in defining what is considered 'tidy', introducing an additional layer of complexity\cite{kant2022housekeep,kapelyukh2022my,zhao2023differentiable,batra2020rearrangement,garrett2021integrated,lin2022survey}. For a robot to be effective in such a setting, it must not only detect and recognize objects but also understand the context in which they are placed and align its actions with human aesthetic and organizational preferences.

Recent studies have drawn parallels from the field of natural language processing, conceptualizing objects on a table as words and the organized table itself as a sentence. Just as words can be rearranged in multiple ways to convey the same meaning in a sentence, objects can be organized in various configurations, all of which can be considered tidy\cite{brown2020language}. When employing traditional supervised learning, which relies on one-to-one mapping, to address multi-label learning, it frequently produces an output that averages the labels.\cite{gibaja2014multi}. Given the multifaceted nature of tidiness, where multiple configurations can be valid, we use an auto-regression transformer-based model to predict the target position of objects\cite{vaswani2017attention}. While such methods have shown promise in knolling tasks, challenges arise when objects are closely packed or stacked, a common scenario in real-world settings\cite{chang2012interactive,katz2014perceiving,katz2013clearing,pan2022algorithms,wada2022safepicking}.

Self-supervised learning allows the robot to generate its supervisory data automatically. It offers a way to harness vast amounts of unlabeled data, allowing robots to learn from interactions with their environment without the need for explicit human supervision\cite{chen2021smile, lee2021sample,pinto2016supersizing,ebert2018robustness,zeng2020tossingbot,fang2020learning}. For the knolling bot, the self-supervised learning paradigm is particularly beneficial. By interacting with objects on a table, the robot can generate a rich dataset that captures the objects and environment dynamics.

In this study, a key contribution is proposing a graspability estimation model, developed through a self-supervised learning framework. This model leverages the output of a visual perception model, equipping the robot with the capability to assess the graspability of objects on a table. If an object is predicted to be ungraspable due to its proximity to others or its position in a stack, the robot arm executes separating behaviors on the object, ensuring a successful grasp. This research aims to develop a robotic system adept at table organization. The methodologies introduced in this work are versatile for potential adaptation to a broader spectrum of domestic tasks. Given the multifaceted challenges of domestic environments, ranging from various object types to diverse spatial configurations, knolling bot 2.0 can handle an arbitrary number of objects, irrespective of their color or size. Future endeavors could explore the scalability of this system, potentially extending its capabilities to larger spaces like rooms or even entire households.

\begin{figure}
\label{method}
\label{method}
  \centering
  \includegraphics[width=0.8\textwidth]{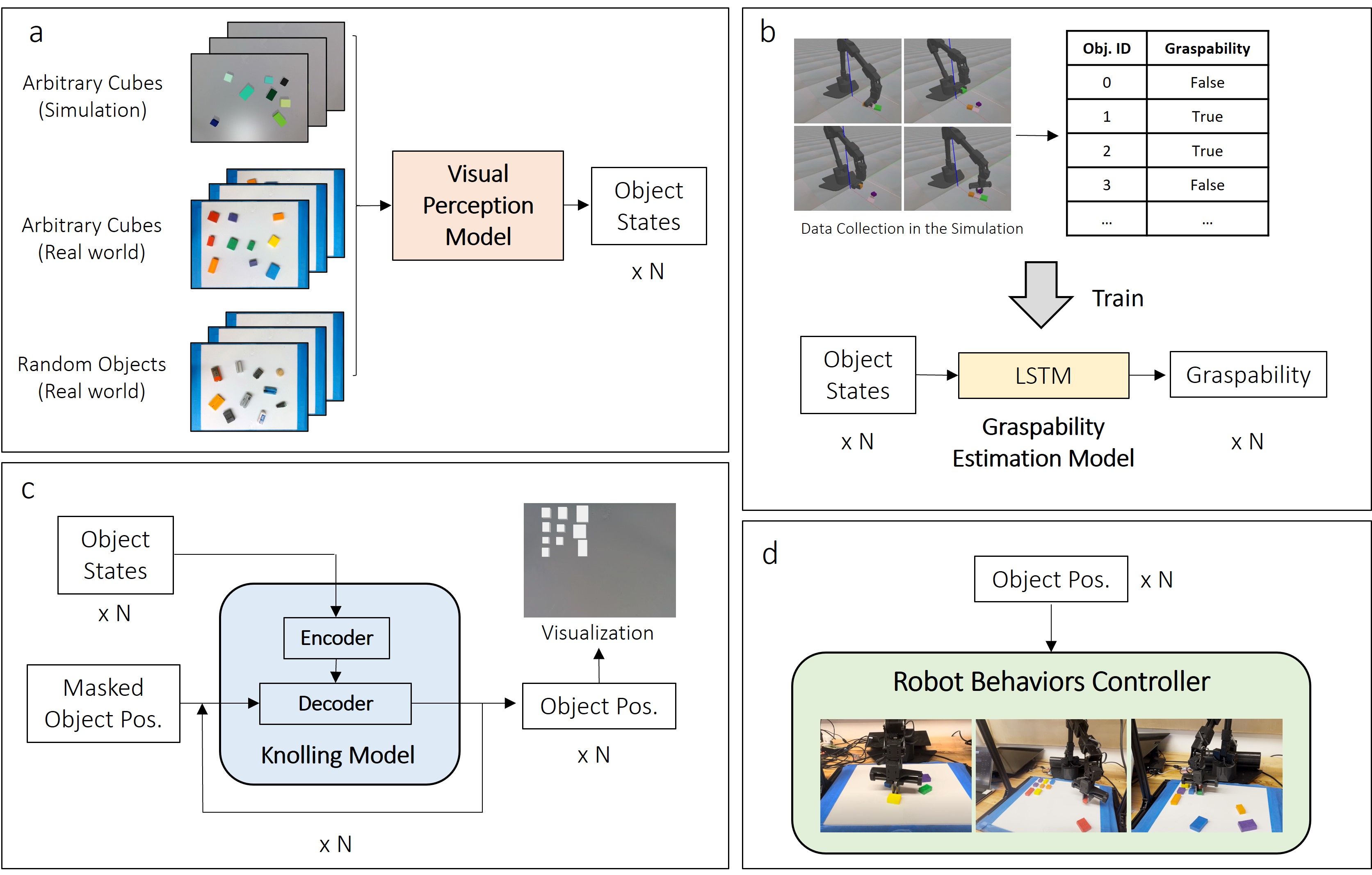}
  \caption{\textbf{Overview of the Knolling Bot 2.0 modules:} \textit{(a)} Training process of the visual perception model using three datasets. \textit{(b)} Self-supervised learning in simulation for training the Graspability Estimation Model (GEM). \textit{(c)} Transformer-based knolling model predicting target positions for objects. \textit{(d)} Robot behaviors controller executing separation and pick-and-place tasks."}
\end{figure}

\section{METHOD}
The proposed Knolling Bot 2.0 is a comprehensive system that integrates four primary modules, each designed to address specific challenges in the knolling process.

\paragraph{Visual Perception Model}
The visual perception model is based on a customized YOLO v8 architecture \cite{reis2023real}. As illustrated in Figure \ref{method}\textit{a}, the training process for this model involves three datasets. The first dataset is derived from simulations using arbitrary cubes. The second dataset comprises real-world images of arbitrary cubes. The third dataset is more diverse, containing images of everyday objects such as batteries, electronics, and erasers. The model processes a single (480, 640) RGB image and produces an (N, 4, 2) format dataset, where N represents the number of objects detected. Each object is associated with four key points, each having two coordinates (x, y). To bridge the simulation-to-reality gap, a visual domain randomization technique is applied \cite{tremblay2018training}, introducing variations in brightness, ground textures, and object appearances.

\paragraph{Graspability Estimation Model (GEM)}
As illustrated in Figure \ref{method}\textit{b}, the self-supervised learning process used to train the GEM involves a robot arm operating within a simulation environment (Movie S2). In the simulation, we initialize an arbitrary number of objects, up to a maximum of five, in two distinct scenarios: crowded and sparse object arrangements. The robot arm attempts to grasp each object, and upon a successful grasp, the object is removed, creating a new environment configuration. This dynamic adjustment ensures sample efficiency as the robot continues grasping in the updated environment, forming a new data pair. This self-generated dataset is then employed to train the GEM for predicting object graspability.

The model is designed to handle a maximum input length of 5 objects, with each object represented by 6 features: its location in x, y, yaw, width, length, and a prediction confidence provided by the visual perception model. When the environment contains more than five objects, the object state list is partitioned into multiple data pairs for processing. GEM is built around a bidirectional LSTM network comprising 8 layers and 32 hidden units, with an input size of 6. Activation functions, including ReLU, are applied to introduce non-linearity, while a Softmax function ensures the output probabilities are normalized. The model employs CrossEntropyLoss as its loss function, optimizing it for classification tasks.

\paragraph{Knolling Model}
The knolling model, depicted in Figure \ref{method}\textit{c}, employs a transformer architecture, designed to predict the optimal placement of objects on a table. The encoder takes a list of object state information, such as width and length. The autoregressive nature of the transformer model is leveraged to predict the target position of objects iteratively. The model training involves two phases: pre-training and fine-tuning. Pre-training focuses on simpler tasks with fewer objects, while fine-tuning emphasizes full knolling tasks. Gaussian mixture models (GMM) are used to handle the inherent variability in object placements.

\paragraph{Robot Arm Controller}
The robot arm controller, as visualized in Figure \ref{method}\textit{d}, synthesizes the outputs from the previous models to execute pick and place tasks. It operates in several modes, including movement between locations, grasping and releasing objects, and object separation. The knolling model provides target positions, while the visual perception model offers current positions. Cartesian Control is employed to ensure smooth trajectories, and the arm can do sweeping or separation behaviors as required based on the predictions of the grasp-predicting model.

\section{Experiments}

To evaluate the effectiveness of the Graspability Estimation Model (GEM), we conducted quantitative experiments comparing it against a baseline method. The baseline relies solely on the prediction confidence from the visual perception model to determine the graspability of objects. This method assumes that stacked objects, when viewed from a top-down camera, exhibit specific shapes, leading to discrepancies between the predicted and actual dimensions of the objects. Thus, the baseline uses prediction confidence as a heuristic to decide whether the objects can be grasped or not.

\begin{figure}[h]
\label{fig_confusion_matrix}
  \centering
  \includegraphics[width=0.8\textwidth]{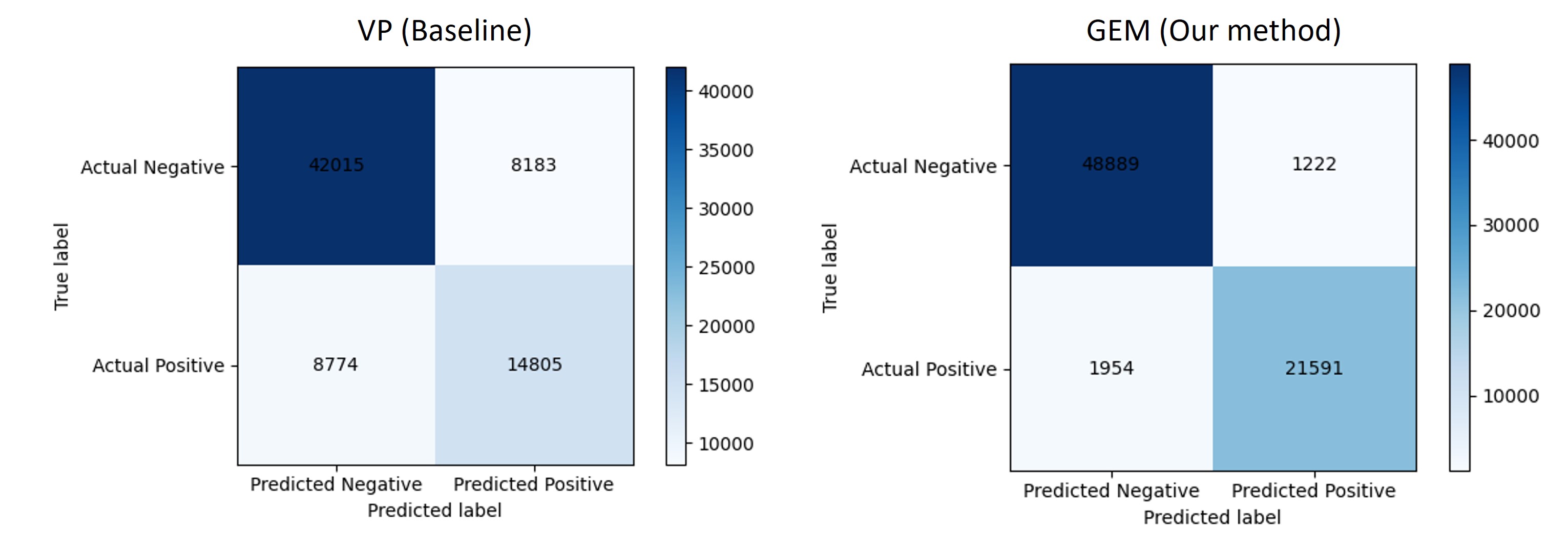}
  \caption{Confusion matrices comparing the baseline Visual Perception (VP) method (left) and our Graspability Estimation Model (GEM) method (right). The matrices provide a detailed breakdown of true positives (TP), true negatives (TN), false positives (FP), and false negatives (FN) for both methods, highlighting the improved accuracy and precision of the GEM}
\end{figure}

For our experiments, we utilized 20,000 data samples from the validation dataset, with each sample containing 2 to 5 objects. This resulted in a total of 73,777 objects. To ensure a fair comparison, we treated the threshold as a continuous variable, using an interval of 0.02. We then selected the threshold value for the baseline that yielded the highest accuracy. Both methods' prediction errors and accuracy results are presented in Table 1. We observe a significant improvement in accuracy when using the GEM over the baseline VP method. The mean prediction error for the GEM is 0.13478, which is considerably lower than the 0.95588 error of the VP method. The GEM achieves an accuracy of 95.688\%, which is markedly higher than the 77.015\% accuracy of the VP method. This substantial increase in accuracy demonstrates the effectiveness of the GEM in predicting object graspability. In the confusion matrix (Figure \ref{fig_confusion_matrix}), the GEM shows better performance in terms of accuracy and precision when compared to the baseline VP method. While the GEM tends to be more conservative in its predictions, leading to a higher FN rate, its significantly reduced FP rate ensures that it is highly likely to be accurate when it predicts an object as graspable. This makes the GEM a more reliable model for real-world robotic applications where precision in grasp predictions is crucial.

\begin{table}[h]
\label{tab_1}
\caption{Prediction Errors and Accuracy}
\centering
\begin{tabular}{|c|c|c|c|c|c|}
\hline
              & Mean    & Std     & Min     & Max     & Accuracy \\ \hline
VP (Baseline) & 0.95588 & 0.02954 & 0.53874 & 0.99324 & 77.015\%  \\ \hline
GEM (OM)      & 0.13478 & 0.10058 & 0.11484 & 0.36035 & \textbf{95.688\%}  \\ \hline
\end{tabular}
\end{table}

\section{Conclusion}
In this paper, we introduced Knolling Bot 2.0, a robotic system that integrates visual perception model, graspability model, knolling model, and robot arm controllers to achieve an arbitrary number of objects organization. The key contribution in this iteration is incorporating the Graspability Estimation Model (GEM), which empowers the robot to adeptly handle and organize piles of objects, a common challenge in real-world settings. In the future, by leveraging DRL, we anticipate achieving more refined control for downstream tasks, especially in scenarios that require intricate object separation and closed-loop pick-and-place operations. 


\bibliography{ref}

\begin{thebibliography}{10}

\bibitem{kapelyukh2023dall}
I.~Kapelyukh, V.~Vosylius, and E.~Johns, ``Dall-e-bot: Introducing web-scale diffusion models to robotics,'' {\em IEEE Robotics and Automation Letters}, 2023.

\bibitem{liu2022structformer}
W.~Liu, C.~Paxton, T.~Hermans, and D.~Fox, ``Structformer: Learning spatial structure for language-guided semantic rearrangement of novel objects,'' in {\em 2022 International Conference on Robotics and Automation (ICRA)}, pp.~6322--6329, IEEE, 2022.

\bibitem{xu2023tidy}
Y.~Xu and D.~Hsu, ``How to tidy up a table: Fusing visual and semantic commonsense reasoning for robotic tasks with vague objectives,'' {\em arXiv preprint arXiv:2307.11319}, 2023.

\bibitem{zeng2021transporter}
A.~Zeng, P.~Florence, J.~Tompson, S.~Welker, J.~Chien, M.~Attarian, T.~Armstrong, I.~Krasin, D.~Duong, V.~Sindhwani, {\em et~al.}, ``Transporter networks: Rearranging the visual world for robotic manipulation,'' in {\em Conference on Robot Learning}, pp.~726--747, PMLR, 2021.

\bibitem{manuelli2019kpam}
L.~Manuelli, W.~Gao, P.~Florence, and R.~Tedrake, ``kpam: Keypoint affordances for category-level robotic manipulation,'' in {\em The International Symposium of Robotics Research}, pp.~132--157, Springer, 2019.

\bibitem{goodwin2022semantically}
W.~Goodwin, S.~Vaze, I.~Havoutis, and I.~Posner, ``Semantically grounded object matching for robust robotic scene rearrangement,'' in {\em 2022 International Conference on Robotics and Automation (ICRA)}, pp.~11138--11144, IEEE, 2022.

\bibitem{danielczuk2021object}
M.~Danielczuk, A.~Mousavian, C.~Eppner, and D.~Fox, ``Object rearrangement using learned implicit collision functions,'' in {\em 2021 IEEE International Conference on Robotics and Automation (ICRA)}, pp.~6010--6017, IEEE, 2021.

\bibitem{lee2021service}
I.~Lee, ``Service robots: A systematic literature review,'' {\em Electronics}, vol.~10, no.~21, p.~2658, 2021.

\bibitem{kim2019control}
J.~Kim, A.~K. Mishra, R.~Limosani, M.~Scafuro, N.~Cauli, J.~Santos-Victor, B.~Mazzolai, and F.~Cavallo, ``Control strategies for cleaning robots in domestic applications: A comprehensive review,'' {\em International Journal of Advanced Robotic Systems}, vol.~16, no.~4, p.~1729881419857432, 2019.

\bibitem{abdo2015robot}
N.~Abdo, C.~Stachniss, L.~Spinello, and W.~Burgard, ``Robot, organize my shelves! tidying up objects by predicting user preferences,'' in {\em 2015 IEEE international conference on robotics and automation (ICRA)}, pp.~1557--1564, IEEE, 2015.

\bibitem{kant2022housekeep}
Y.~Kant, A.~Ramachandran, S.~Yenamandra, I.~Gilitschenski, D.~Batra, A.~Szot, and H.~Agrawal, ``Housekeep: Tidying virtual households using commonsense reasoning,'' in {\em European Conference on Computer Vision}, pp.~355--373, Springer, 2022.

\bibitem{kapelyukh2022my}
I.~Kapelyukh and E.~Johns, ``My house, my rules: Learning tidying preferences with graph neural networks,'' in {\em Conference on Robot Learning}, pp.~740--749, PMLR, 2022.

\bibitem{zhao2023differentiable}
Z.~Zhao, W.~S. Lee, and D.~Hsu, ``Differentiable parsing and visual grounding of natural language instructions for object placement,'' in {\em 2023 IEEE International Conference on Robotics and Automation (ICRA)}, pp.~11546--11553, IEEE, 2023.

\bibitem{batra2020rearrangement}
D.~Batra, A.~X. Chang, S.~Chernova, A.~J. Davison, J.~Deng, V.~Koltun, S.~Levine, J.~Malik, I.~Mordatch, R.~Mottaghi, {\em et~al.}, ``Rearrangement: A challenge for embodied ai,'' {\em arXiv preprint arXiv:2011.01975}, 2020.

\bibitem{garrett2021integrated}
C.~R. Garrett, R.~Chitnis, R.~Holladay, B.~Kim, T.~Silver, L.~P. Kaelbling, and T.~Lozano-P{\'e}rez, ``Integrated task and motion planning,'' {\em Annual review of control, robotics, and autonomous systems}, vol.~4, pp.~265--293, 2021.

\bibitem{lin2022survey}
T.~Lin, Y.~Wang, X.~Liu, and X.~Qiu, ``A survey of transformers,'' {\em AI Open}, 2022.

\bibitem{brown2020language}
T.~Brown, B.~Mann, N.~Ryder, M.~Subbiah, J.~D. Kaplan, P.~Dhariwal, A.~Neelakantan, P.~Shyam, G.~Sastry, A.~Askell, {\em et~al.}, ``Language models are few-shot learners,'' {\em Advances in neural information processing systems}, vol.~33, pp.~1877--1901, 2020.

\bibitem{gibaja2014multi}
E.~Gibaja and S.~Ventura, ``Multi-label learning: a review of the state of the art and ongoing research,'' {\em Wiley Interdisciplinary Reviews: Data Mining and Knowledge Discovery}, vol.~4, no.~6, pp.~411--444, 2014.

\bibitem{vaswani2017attention}
A.~Vaswani, N.~Shazeer, N.~Parmar, J.~Uszkoreit, L.~Jones, A.~N. Gomez, {\L}.~Kaiser, and I.~Polosukhin, ``Attention is all you need,'' {\em Advances in neural information processing systems}, vol.~30, 2017.

\bibitem{chang2012interactive}
L.~Chang, J.~R. Smith, and D.~Fox, ``Interactive singulation of objects from a pile,'' in {\em 2012 IEEE International Conference on Robotics and Automation}, pp.~3875--3882, IEEE, 2012.

\bibitem{katz2014perceiving}
D.~Katz, A.~Venkatraman, M.~Kazemi, J.~A. Bagnell, and A.~Stentz, ``Perceiving, learning, and exploiting object affordances for autonomous pile manipulation,'' {\em Autonomous Robots}, vol.~37, pp.~369--382, 2014.

\bibitem{katz2013clearing}
D.~Katz, M.~Kazemi, J.~A. Bagnell, and A.~Stentz, ``Clearing a pile of unknown objects using interactive perception,'' in {\em 2013 IEEE International Conference on Robotics and Automation}, pp.~154--161, IEEE, 2013.

\bibitem{pan2022algorithms}
Z.~Pan, A.~Zeng, Y.~Li, J.~Yu, and K.~Hauser, ``Algorithms and systems for manipulating multiple objects,'' {\em IEEE Transactions on Robotics}, 2022.

\bibitem{wada2022safepicking}
K.~Wada, S.~James, and A.~J. Davison, ``Safepicking: Learning safe object extraction via object-level mapping,'' in {\em 2022 International Conference on Robotics and Automation (ICRA)}, pp.~10202--10208, IEEE, 2022.

\bibitem{chen2021smile}
B.~Chen, Y.~Hu, L.~Li, S.~Cummings, and H.~Lipson, ``Smile like you mean it: Driving animatronic robotic face with learned models,'' in {\em 2021 IEEE International Conference on Robotics and Automation (ICRA)}, pp.~2739--2746, IEEE, 2021.

\bibitem{lee2021sample}
R.~Lee, M.~Hamaya, T.~Murooka, Y.~Ijiri, and P.~Corke, ``Sample-efficient learning of deformable linear object manipulation in the real world through self-supervision,'' {\em IEEE Robotics and Automation Letters}, vol.~7, no.~1, pp.~573--580, 2021.

\bibitem{pinto2016supersizing}
L.~Pinto and A.~Gupta, ``Supersizing self-supervision: Learning to grasp from 50k tries and 700 robot hours,'' in {\em 2016 IEEE international conference on robotics and automation (ICRA)}, pp.~3406--3413, IEEE, 2016.

\bibitem{ebert2018robustness}
F.~Ebert, S.~Dasari, A.~X. Lee, S.~Levine, and C.~Finn, ``Robustness via retrying: Closed-loop robotic manipulation with self-supervised learning,'' in {\em Conference on robot learning}, pp.~983--993, PMLR, 2018.

\bibitem{zeng2020tossingbot}
A.~Zeng, S.~Song, J.~Lee, A.~Rodriguez, and T.~Funkhouser, ``Tossingbot: Learning to throw arbitrary objects with residual physics,'' {\em IEEE Transactions on Robotics}, vol.~36, no.~4, pp.~1307--1319, 2020.

\bibitem{fang2020learning}
K.~Fang, Y.~Zhu, A.~Garg, A.~Kurenkov, V.~Mehta, L.~Fei-Fei, and S.~Savarese, ``Learning task-oriented grasping for tool manipulation from simulated self-supervision,'' {\em The International Journal of Robotics Research}, vol.~39, no.~2-3, pp.~202--216, 2020.

\bibitem{reis2023real}
D.~Reis, J.~Kupec, J.~Hong, and A.~Daoudi, ``Real-time flying object detection with yolov8,'' {\em arXiv preprint arXiv:2305.09972}, 2023.

\bibitem{tremblay2018training}
J.~Tremblay, A.~Prakash, D.~Acuna, M.~Brophy, V.~Jampani, C.~Anil, T.~To, E.~Cameracci, S.~Boochoon, and S.~Birchfield, ``Training deep networks with synthetic data: Bridging the reality gap by domain randomization,'' in {\em Proceedings of the IEEE conference on computer vision and pattern recognition workshops}, pp.~969--977, 2018.

\end{thebibliography}
\end{document}